\pdfoutput=1

\documentclass[11pt]{article}

\usepackage[final]{acl}

\usepackage{times}
\usepackage{latexsym}
\usepackage[T1]{fontenc}
\usepackage[utf8]{inputenc}
\usepackage{times}
\usepackage{latexsym}
\usepackage[inline]{enumitem}

\usepackage[T1]{fontenc}

\usepackage[utf8]{inputenc}

\usepackage{microtype}
\usepackage{graphicx}
\usepackage{multirow}
\usepackage{booktabs}

\usepackage{acronym}

\usepackage{tabularx}
\usepackage{algorithm}
\usepackage{algpseudocode}
\usepackage{subcaption}
\usepackage{array}
\usepackage{makecell}
\usepackage{balance}
\usepackage{lipsum} 

\usepackage[skip=2pt]{caption}
\newcommand{\header}[1]{\vspace*{1mm}\noindent\textbf{#1}}

\acrodef{QA}{question answering}
\acrodef{MC}{machine comprehension}

\acrodef{PeaQA}{parameter-efficient abstractive question answering}

\usepackage{xcolor}
\usepackage{array, makecell}
\usepackage[utf8]{inputenc}
\usepackage{enumitem}
\usepackage{adjustbox}
\usepackage{longtable}
\usepackage{multicol}
\usepackage{supertabular}
\title{Parameter-Efficient Abstractive Question Answering\\ over Tables or Text}

\author{
Vaishali Pal$^1$
\qquad
Evangelos Kanoulas$^2$
\qquad
Maarten de Rijke$^2$
\\
$^1$Discovery Lab, University of Amsterdam
\qquad
$^2$University of Amsterdam \\
\texttt{v.pal, e.kanoulas, m.derijke@uva.nl}
}

\begin{document}
\maketitle
\begin{abstract}
A long-term ambition of information seeking \ac{QA} systems is to reason over multi-modal contexts and generate natural answers to user queries. Today, memory intensive pre-trained language models are adapted to downstream tasks such as \ac{QA} by fine-tuning the model on \ac{QA} data in a specific modality like unstructured text or structured tables. To avoid training such memory-hungry models while utilizing a uniform architecture for each modality, parameter-efficient adapters add and train small task-specific bottle-neck layers between transformer layers. In this work, we study parameter-efficient abstractive \ac{QA} in encoder-decoder models over structured tabular data and unstructured textual data using only 1.5\% additional parameters for each modality. We also ablate over adapter layers in both encoder and decoder modules to study the efficiency-performance trade-off and demonstrate that reducing additional trainable parameters down to 0.7\%--1.0\% leads to comparable results. Our models out-perform current state-of-the-art models on tabular QA datasets such as Tablesum and FeTaQA, and achieve comparable performance on a textual QA dataset such as NarrativeQA using significantly less trainable parameters than fine-tuning.
\end{abstract}

\acresetall

\section{Introduction}
Information seeking systems over diverse contexts require model capabilities to reason over unstructured and structured 
 \begin{figure}[ht!]
 \centering
    \includegraphics[width=\columnwidth]{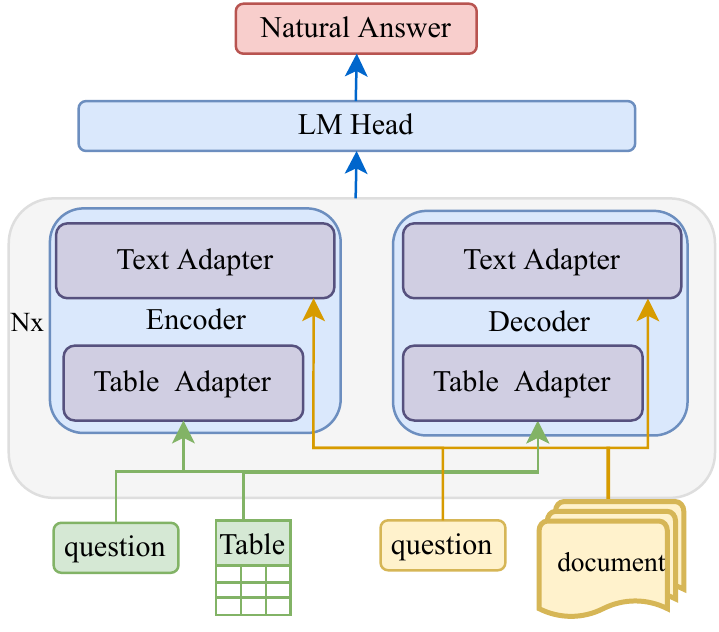}
 \caption{Parameter-efficient transfer learning using modality-specific (table/text) adapters for Abstractive Question Answering}
 \label{fig:architecture}
\end{figure}
data such as free-form text, tables, and images \citep{agrawal2016vqa,vakulenko2018qrfa,hudson2019gqa,Zhang:2020:SET,zhu2021tatqa,DBLP:conf/sigir/DeldjooTZ21}. Such systems might have the additional requirement of generating natural language responses if deployed as task-oriented conversational agents \citep{wen2015semantically,Carnegie00stochasticlanguage,rambow-etal-2001-natural,RATNAPARKHI2002435}. 
Recent work on open-domain \ac{QA} predominately addresses these challenges by fine-tuning massive pre-trained language models on different modalities such as tables and text  \citep{tabert2020,herzig-etal-2020-tapas,2021,katsis2021aitqa,nan2021feta}. However, each model trained on a specific input type is incompatible with other modalities and requires modality-specific fine-tuning. 
For example, in tabular \ac{QA} \citep{herzig-etal-2020-tapas}, the structure of the table is learnt by training additional position embeddings (row and column identifiers) to identify which row and column a table cell belongs to. This renders such modality specific models incompatible with free-form text-based models. Multi-modal models \citep{zhu2021tatqa} can reason over both tables and text by concatenating the textual context and the flattened table, leading to longer input sequences and limiting the length of the context that can be encoded. 

To address these challenges, we study parameter-efficient transfer learning for abstractive \ac{QA} over tables and over text.
We are motivated to use adapter-layers that inject small bottle-neck layers between frozen pre-trained transformer layers as they achieve comparable performance to fine-tuning on a variety of tasks such as multi-lingual translation \citep{pfeiffer-etal-2020-mad, philip-etal-2020-monolingual,NEURIPS2020_7a6a74cb}, classification \citep{pmlr-v97-houlsby19a}, text-to-text generation \citep{lin-etal-2020-exploring}, domain-adaptation in dialogue state tracking, and response generation \citep{hung2021dstod}. 

Ablation studies on adapter layers \citep{ruckle2020adapterdrop} on masked language models such as BERT-base and RoBERTa over the GLUE benchmark demonstrate that removing beginning adapter layers leads to a minimal drop in performance. Extending adapter layer ablation over separate encoder and decoder modules is non-trivial as the conventional approach of sequential pruning of layers does not extend to consecutive encoder and decoder modules. Our work explores the interaction of adapter layers from both modules in the context of abstractive QA.

\citet{lin-etal-2020-exploring} explore the impact of the adapter bottle-neck dimension for various language generation tasks over an auto-regressive model such as GPT-2 \citep{radford2019language}. They do not study tabular data nor ablate adapter layers, which is crucial in understanding impact of individual adapters in sequential transformer module architectures such as encoder-decoder. Our analysis is complementary to \citep{lin-etal-2020-exploring} as we ablate adapter layers to study parameter-performance trade-off whereas they only focus on adapter bottle-neck size. Also, we generalize beyond the text-to-text setting and explore language generation from structured or unstructured input such as tables and text. This introduces domain-shift in both the \emph{task} and \emph{structure} of the downstream data.

We propose a system, named \textbf{P}arameter, \textbf{E}fficient, \textbf{A}bstractive \textbf{Q}uestion \textbf{A}nswering (PeaQA), shown in Figure \ref{fig:architecture}, which learns to reason over unstructured and structured input using a \emph{shared} pre-trained language model and modality-specific adapter layers. We  automatically transform hierarchical tables to regular tables to have a uniform representation without breaking associations between table cells. In addition, we extend the study of ablating adapter layers over both encoder and decoder modules. 

Our main contributions are summarized as:
\begin{enumerate}[label=(\arabic*),leftmargin=*,nosep]
    \item We perform parameter-efficient abstractive question answering over multi-modal context using only additional 1.5\% of trainable parameters for each modality. Our adapter-tuned model outperforms existing work by a large margin on tabular QA datasets and achieves comparable performance on a textual QA dataset. 
    \item We study tabular QA as a new modality that introduces massive input domain shift to pre-trained language models. We propose a 2-step transformation of hierarchical tables to sequences, which produces a uniform representation to be used by a single, shared pre-trained language model and  modality-specific adapter layers. To the best of our knowledge, this is the first work that explores tabular QA question answering in a parameter-efficient manner.
    \item We ablate adapter layers in both encoder and decoder modules to study their impact and show that beginning layers from both encoder and decoder can be eliminated without significant drop in performance. We also demonstrate that last encoder adapter layers are indispensable and have greater contribution than decoder layers at the same level.
\end{enumerate}

\section{Related Work}
\textbf{Tabular question answering.} 
Tabular \ac{QA} systems aim to answer questions from structured tables, which can be regular or hierarchical. Hierarchical tables can have header cells and body cells spanning across multiple rows and columns \citep{Cheng2021HiTabAH}. In most tabular QA systems \citep{herzig-etal-2020-tapas,zhu2021tatqa,katsis2021aitqa}, the structure of the table is  encoded in the embedding layer of large language models by introducing table specific position information such as row id and column id. Concurrent to our work, abstractive \ac{QA} over tables \citep{nan2021feta,Cheng2021HiTabAH} poses additional challenges of generating natural answers by reasoning and aggregating discontinuous facts from the table. 

\header{Textual question answering.} Question answering over text measures a system's ability to comprehend free-form text in the user question and context passage(s) and predict an answer. The answer predicted can be extractive in nature, where the system identifies short text spans in the context passage to answer the user query \citep{lee2016learning,seo2016bidirectional,2016_squad,pearce2021comparative}, or it can be abstractive, where it is required to generate a free-form answer \citep{2016neu_gen,mitra2017generative,Bauer2018CommonsenseFG,reddy2019coqa}.

\header{Transfer learning.}
Transfer learning techniques such as fine-tuning pre-trained models for downstream tasks, require a new set of parameters to be learnt for each new task. To avoid such memory intensive transfer learning methods, adapters have been proposed as a parameter-efficient method of adapting to new domains~\citep{DBLP:journals/corr/abs-1902-00751,pfeiffer-etal-2020-mad}. Adapters have been extended to language generation in a variety of generative tasks such as translation, summarization, multi-turn dialogue, and task-oriented natural language generation~\citep{lin-etal-2020-exploring}. 

Our work combines all the aforementioned aspects to generate abstractive answers from \emph{both} tables and text with only $0.7\%$--1.0$\%$ trainable parameters without compromising performance.

\section{Model}
We focus on encoder-decoder models for the task of abstractive question answering. We use a BART \citep{lewis2019bart} encoder-decoder architecture which comprises of a bidirectional encoder and an auto-regressive decoder. The input sequence consists of the question, the context title and context sequence preceded with prompts indicating the beginning of the each sub-sequence. Formally, the input sequence is represented as ${<}$\emph{question}${>}$ $q_0$  $q_1$  \ldots\  $q_m$ ${<}$\emph{title}${>}$ $t_1$ $t_2$ \ldots\ $t_p$ ${<}$\emph{context}${>}$ $c_0$ $c_1$ \ldots\ $c_n$, where $q_i$ is the $i$-th question token, $t_j$ is the $j$-th title token, and $c_k$ is the $k$-th context token. The context can either be a text passage or a flattened table. The parameters of the pre-trained BART model are frozen during training. Modality specific adapter layers added to the model are trained on either tabular context or textual context to generate natural answers.

\section{Textual Question Answering} 
To study multi-modal abstractive QA, we first focus on free-form text as context to the system. We train adapter layers for textual context on the NarrativeQA dataset \citep{kocisky2017narrativeqa}. NarrativeQA is a complex abstractive question answering dataset over stories.  The dataset contains $32,747$ samples in the training set, $3,461$ samples in the validation set, and $10,557$ samples in the test set. For our task, we have selected the input context passage to be the human annotated \emph{summary} of each sample which is the Wikipedia page summary of the story and represented as a paragraph. The input to the model is the \emph{question}, \emph{title} and \emph{summary} of each passage and the target is the abstractive answer. 

\section{Tabular Question Answering} 
\label{sec:Tab_QA}
We study tabular QA as a new modality which introduces massive input domain shift to pre-trained language models. Tables enforce structural constraints in their representation which is incompatible with the expected input format of pre-trained language models. To achieve our goal of parameter efficiency by utilizing a uniform pre-trained language model, we only train table specific adapter layers while keeping the pre-trained model frozen. However, this necessitates a uniform input representation for both tables and text. An additional challenge is introduced to maintain uniformity across different table types (regular, hierarchical).   
 
For our task, we explore $2$ tabular QA datasets, namely, Tablesum \citep{Zhang:2020:SET} and FeTaQA \citep{nan2021feta}. Tablesum consists of 200 unique Wikipedia tables over which questions and abstractive answers are manually annotated; 40\% of the samples are questions over hierarchical tables but the tables in their released data are missing information in the hierarchical cells and their work do not handle hierarchies. We address this issue by extracting the wikitables from the respective Wikepedia pages and release a clean version of the dataset.\footnote{The cleaned data and code can be found at \url{ https://github.com/kolk/Pea-QA}} 
 
 FeTaQA \citep{nan2021feta} is a larger abstractive tabular QA dataset consisting of question and free-form answers over $10,330$ regular tables. The dataset consists of $7,326$ samples in the training set, $1,001$ in the validation set, and $2,003$ in the test set. FeTaQA consists of human-annotated answers containing explanations involving entities and relations. 
 
\subsection{Table Representation}
For our work, we choose to represent all tables uniformly in a two-step process:
\begin{enumerate*}[label=(\arabic*)]
\item Transformation of a hierarchical table into a regular table; and 
\item Linearization of a regular table into a flattened sequence which can be encoded with a language model.
\end{enumerate*}

\begin{figure}
\begin{subfigure}{.5\textwidth}
  \centering
   \includegraphics[scale=0.4]{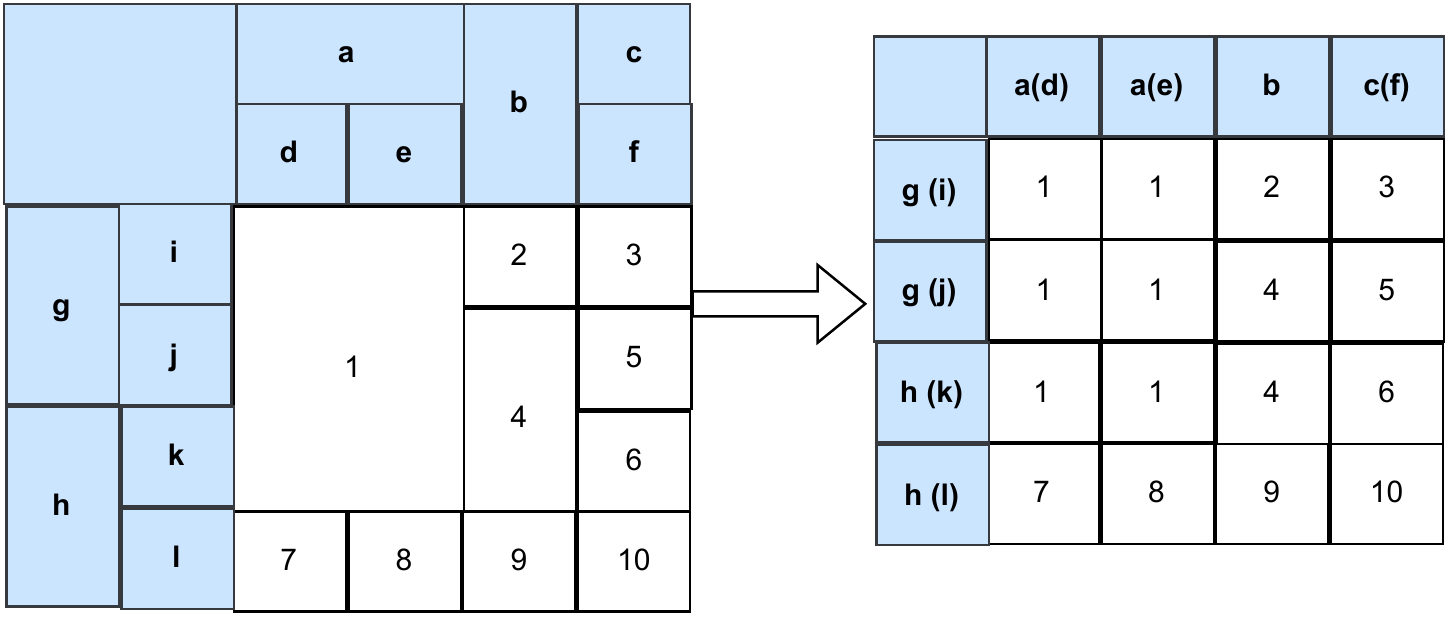}
 \caption{A multi-span table represented as a regular table.}
 \label{fig:multispan2regulartable}
\end{subfigure}
\newline
\begin{subfigure}{.5\textwidth}
  \centering
  \includegraphics[scale=0.4]{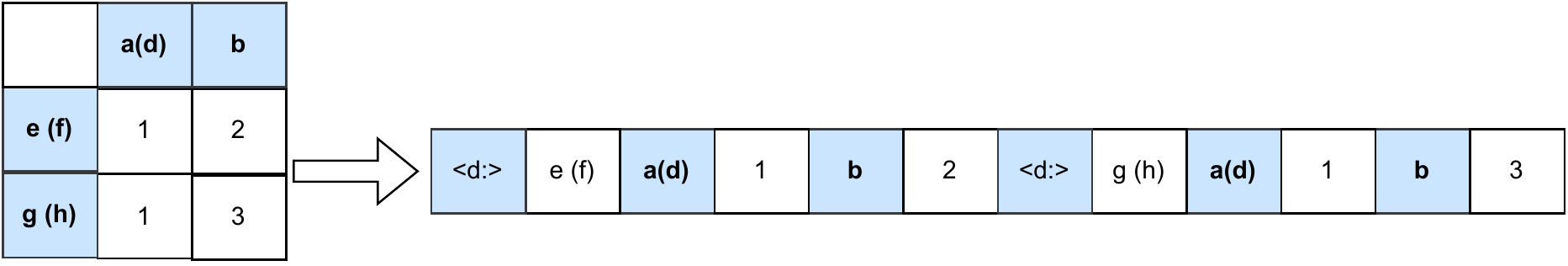}
 \caption{Linearize regular table to a sequence of \emph{key}:\emph{value} pairs.}
 \label{fig:regulartable2sequence}
\end{subfigure}
\caption{Table representation.}
\label{fig:fig}
\end{figure}

\header{Linearize hierarchical table headers.}  
Hierarchical table headers are linearized into a single row of headers by the following process.  A header cell spanning multiple columns is duplicated and split into multiple cells. Next, the cell values over which this header spans are concatenated with the entire split.  Repeating this process over all header rows flattens the hierarchical header into a sequential one. We depict this process in Figure~\ref{fig:multispan2regulartable}, which yields a linear header $a(d)$, $a(d)$, $b$, $e(f)$.

\header{Linearizing table body.} 
Multi-span table body cells are parsed differently than headers. Each table body cell is replicated with one or multiple header cells depending on its span across columns. Cells that span across multiple rows are replicated with all the spanned rows. This process leads to a regular table. We flatten the regular table in row-major form, concatenating rows sequentially. Each row is a sequence of \emph{(key, value)} pairs where a key is a column header and the value is the cell value of that column as depicted in Figure \ref{fig:regulartable2sequence}.

\section{Experimental Setup}
\label{sec:experimental_setup}
We seek to answer the following research questions with our experiments:
\begin{enumerate*}[label=(RQ\arabic*)]
\item How does adapter-tuning perform compared to fine-tuning in the context of multi-modal input?
\item Do all adapter layers across the encoder and decoder contribute equally to performance across tasks/modalities? 
\end{enumerate*}

\subsection{Fine-Tuning}
\label{sec:experiment_finetuning}
We perform all our experiments on the \emph{large} variant of BART model. We fine-tune the BART-large model over the 3 datasets as the state-of-the-art fine-tuned models utilize different architectures for different datasets making comparison with adapter-tuning difficult. We treat our fine-tuned BART models on the 3 datasets as baselines. We sweep learning rates from \{$8e^{-4}$, $6e^{-4}$, $3e^{-4}$, $1e^{-4}$, $5e^{-5}$, $4^e{-5}$, $3e^{-5}$, $2e^{-5}$, $1e^{-5}$\} and select the best performing learning rate for each dataset. We select $4e^{-5}$ for fine-tuning on Tablesum, $8e^{-4}$ on FeTaQA datasets and $2e^{-5}$ to fine-tune NarrativeQA. We use a batch size of $4$ and gradient accumulation of $8$ to emulate an effective batch size of $32$. The maximum target sequence length is set to $200$ for tabular QA datasets and to $100$ for the textual QA dataset. On the Tablesum dataset, we follow  5-fold cross validation as described in the original work to evaluate our models. On FeTaQA and NarrativeQA, we utilize the test split for evaluating our models. We train the model on each dataset for $15$ epochs and evaluate on Rouge-2, Rouge-L and sacreBLEU metrics. 

\subsection{Adapter-Tuning}
We perform adapter-tuning as a parameter-efficient alternative to adapt BART-large model to the abstractive question answering task across different modalities. We first freeze all layers of the pre-trained BART-large model which was trained on text reconstruction as mentioned in the original BART paper \cite{lewis2019bart}. We add bottle-neck adapter layers from the Houlsby adapter configuration \citep{pmlr-v97-houlsby19a} which are trained to adapt to the downstream abstractive question answering task and also to modality specific input context. Each adapter layer has a bottle-neck embedding size of $64$. As mentioned in Section \ref{sec:experiment_finetuning}, we sweep learning rates and select the best performing learning rate for each dataset. We select $6e^{-4}$ for the tabular QA datasets Tablesum and FeTaQA, and select $1e^{-1}$ to train the textual QA dataset NarrativeQA.  We use the same batch size and maximum target sequence length as fine-tuning for effective comparison. A summary of hyper-parameters are mentioned in Table \ref{tab:hyperparamteres_adapter-tune}.
\begin{table}[tbh!]
\centering
\begin{tabular}{llrr}
\toprule
\textbf{Dataset} & \textbf{Params} & \textbf{ATune} & \textbf{FTune}\\
\midrule
 \multirow{2}{*}{\textbf{All}}  & scheduler & linear & linear \\
 & batch size & 32 & 32 \\
 & seed & 6 & 6 \\
 & max epochs & 15 & 15 \\
 \midrule
 \midrule
 \multirow{2}{*}{\textbf{Tablesum}} & learning rate & 6e-4 & 4e-5 \\
  & input length & 200 & 200 \\
 \midrule
 \multirow{2}{*}{\textbf{FeTaQA}}  & learning rate & 6e-4 & 8e-4 \\
 & input length & 100 & 100 \\
 \midrule
  \multirow{2}{*}{\textbf{NarrativeQA}}  & learning rate & 1e-4 & 2e-5\\
 & input length & 50 & 50 \\
\bottomrule
\end{tabular}
\caption{Hyper-parameters for training. \textbf{ATune} indicates Adapter-tuning, \textbf{FTune} indicates Fine-tuning, \textbf{All} indicates all 3 datasets.}
\label{tab:hyperparamteres_adapter-tune}
\end{table}

 \begin{table*}[t!]
\centering
\resizebox{\textwidth}{!}{%
\begin{tabular}{lll cccc}
\toprule
\textbf{Dataset} & \textbf{Model} & \textbf{Training} & \textbf{Rouge-1} & \textbf{Rouge-2} & \textbf{Rouge-L} & \textbf{BLEU} \\
\midrule
\multirow{5}{*}{\makecell{Tablesum \\ \citep{Zhang:2020:SET}}} & GPT2 & \multirow{2}{*}{fine-tune} & $0.272$ & $0.073$ & $0.200$ & $5.35$ \\
& T5 & & $0.362$ & $0.143$ & $0.276$ & $\textbf{10.43}$ \\
\cmidrule{2-7}
& \multirow{2}{*}{\textbf{Ours (Pea-QA)}} & fine-tune(Baseline) & $\textbf{0.400}$ &	$\textbf{0.186}$	& $\textbf{0.316}$ & $6.30$ \\
 & & Adapter-tune & $0.393$ & $\textbf{0.186}$ & $0.312$ &  $6.75$ \\
\midrule
\multirow{5}{*}{\makecell{FeTaQA \\ \citep{nan2021feta}}} & T5-small & \multirow{3}{*}{fine-tune} &   $0.550$ &  $0.330$  & $0.470$  & $21.60$ \\
& T5-base & & $0.610$  &  $0.390$  &  $0.510$ & $28.14$ \\
& T5-large & & $0.630$   & $0.414$ & $0.530$  & $30.54$ \\
\cmidrule{2-7}
& \multirow{2}{*}{\textbf{Ours (Pea-QA)}} & fine-tune(Baseline) & $0.632$ & $0.415$ &	$0.534$  & $30.81$ \\
& & Adapter-tune & \textbf{0.651} & \textbf{0.436}	& \textbf{0.553} &  \textbf{33.45} \\
\midrule
\multirow{4}{*}{\makecell{NarrativeQA \\ \citep{kocisky2017narrativeqa}}} & \makecell{ Masque\phantom{MasqueMasq} \\\citep{nishida-etal-2019-multi}} & fine-tune  &  -- &	-- & $\textbf{0.547}$  & --  \\
\cmidrule{2-7}
& \multirow{2}{*}{\textbf{Ours (Pea-QA)}} & fine-tune(Baseline) & $0.518$ & $0.268$ & $0.515$ &  $21.07$ \\
 & & Adapter-tune & $0.510$ & $0.270$ & $0.500$ & $20.08$ \\
\bottomrule
\end{tabular}
}
\caption{Results: Scores obtained on the Tablesum, FeTaQA and NarrativeQA datasets.}
\label{tab:tablesum_results}
\end{table*}

\subsection{Ablation Study: Adapter Pruning}
Adapter-layer pruning has been explored on the GLUE benchmark in \citep{ruckle2020adapterdrop}, which demonstrates that removing adapter layers from the beginning of BERT-base and RoBERTa models leads to minimal performance drop. We extend adapter layer ablation to encoder-decoder architectures and hypothesize that this phenomenon should be observed on both the encoder and decoder modules. However, it is non-trivial how the adapter-layers in the encoder and decoder interact with each other and contribute to performance. Previous studies \citep{ruckle2020adapterdrop}  on adapter ablation prune consecutive adapter layers in masked language models. This approach does not extend directly to sequential modules of encoder-decoder where intra-module adapters not only contribute to their respective objective of encoding and decoding but also contributes to inter-module interaction and performance.  To measure the impact of the adapter layers in different modules, we perform adapter ablation in both the encoder and decoder. First, we uniformly remove adapter layers from both encoder and decoder modules starting from the beginning layers of both modules and finally deleting all layers. This leads to 12 experiments corresponding to eliminating 12 encoder and 12 decoder adapter layers. To study interaction across inter-module adapters at different levels, we conduct 36 experiments of different configurations of adapter elimination from the last 6 levels of encoder and decoder. We analyze the performance by each configuration in Section \ref{sec:result_ablation}.

\begin{table*}[th!]
\centering
\begin{tabularx}{\linewidth}{|X|X|}
\hline
\textbf{Question:} What and when were Akhila Kishore's first two films?  \\
\textbf{Target:}  akhila kishore made her debut in the kannada film padhe padhe (2013), and appeared in kathai thiraikathai vasanam iyakkam (2014). \\
\textbf{Table:}
 \begin{adjustbox}{width=0.6\textwidth}
\begin{tabular}{ |c|c|c|c| }
\hline
\textbf{Year} & \textbf{Film} & \textbf{Role} & \textbf{Language}   \\
\hline
2013 & Padhe Padhe & Kanchana & Kannada \\
2014 & Kathai Thiraikathai Vasanam Iyakkam & Daksha & Tamil \\
2015 & Inimey Ippadithaan & Akhila & Tamil \\
 ... & ... & ... & \\
 \hline
 \end{tabular}
 \end{adjustbox}
\\
\textbf{Adaper-tune:}  akhila kishore made her debut in the kannada film padhe padhe (2013) and kathai thiraikathai vasanam iyakkam (2014).\\
\textbf{Fine-tune:}  kathai thiraikathai vasanam iyakkam (2014) and inimey ippadithaan (2015) were kannada films. \\
\hline
\textbf{Question:} Who is the starring actor of Aastik? \\
\textbf{Target:} aastik is a 1956 hindi film starring shahu modak, paro devi and meenakshi. \\
\textbf{Table:} 
 \begin{adjustbox}{width=0.8\textwidth}
\begin{tabular}{ |c|c|c| }
\hline
 \textbf{Title} & \textbf{Director} & \textbf{Cast}  \\
 \hline
 ... & ... & ...   \\
 Aastik &  S. P. Kalla &  Shahu Modak, Paro Devi, Meenakshi, B. M. Vyas, Praveen Paul   \\
 Alam Ara &  Nanubhai Vakil & Daljeet, Chitra, Tiwari, Niranjan Sharma, Minu Mumtaz,...  \\
 ... & ... & ...   \\
 \hline
 \end{tabular}
 \end{adjustbox} \\
\textbf{Adaper-tune:}   aastik is a 1956 bollywood film starring shahu modak. \\
\textbf{Fine-tune:} a directed by s. p. kalla. \\
\hline
\textbf{Question:} What were the three films directed by Yakub and when were they released? \\
\textbf{Target:}  yakub directed three films: sagar ka sher in 1937, uski tamanna in 1939, and, in 1949, aiye. \\
\textbf{Table:} 
 \begin{adjustbox}{width=0.45\textwidth}
\begin{tabular}{|c|c|c|}
\hline
 \textbf{Year} & \textbf{Film} & \textbf{Director}  \\ 
 \hline
 ... & ... & ... \\ 
 1937 & Sagar Ka Sher (Lion of Sagar) & Yakub \\
 ... & ... & ... \\
 1939 & Uski Tamanna (Her Last Desire) & Yakub \\
 ... & ... & ... \\
 1949 & Aiye & Yakub \\ 
 ... & ... & ... \\
 \hline
 \end{tabular}
 \end{adjustbox}
 \\

\textbf{Adaper-tune:}  yakub directed three films: sagar ka sher (lion of sagar) in 1937, uski tamanna (her last desire) in 1939 and aiye in 1949.\\
\textbf{Fine-tune:}   y directed by yakub. \\
 \hline
\end{tabularx}
\caption{Samples where adapter-tune outperforms  fine-tune}
\label{tab:fetaqa_examples}
\end{table*}

\begin{table*}[th!]
\centering
\begin{adjustbox}{width=1.0\textwidth}
\begin{tabularx}{\linewidth}{|X|X|}
\hline
\textbf{Question:} how many times was ed sheeran listed as the performer?  \\
\textbf{Targets:} 
    \setlist{nolistsep}
    \begin{itemize}[noitemsep]
     \item Ed Sheeran was listed as a performer twice in the table documenting the top hits of 2014 in Sweden. Other English-Language top performers included Bruce Springsteen, Sam Smith, and Coldplay, implying that English-Language music has significant success in Sweden. 
    \item  According to the table, in 2014, Ed Sheeran was only listed as the performer one time. It was for the song that he performed that is called ""I See Fire"", which was out in January and February of 2014
 \end{itemize}
\textbf{Table: } 
 \begin{adjustbox}{width=0.7\textwidth}
\begin{tabular}{|c|c|c|c|c|c|}
\hline
\textbf{Week} & \textbf{Date} & \textbf{Song title} & \textbf{Performer} & \textbf{Album title} & \textbf{Performer} \\
\hline
1 & 3 Jan 2014	& \multirow{2}{*}{"Timber"} & \multirow{2}{*}{Pitbull feat. ...} &	\multirow{2}{*}{True} &	\multirow{2}{*}{Avicii} \\
\cline{1-2}
2 & 10 Jan 2014 & & & & \\
\hline
3 & 17 Jan 2014	& \multirow{7}{*}{"I See Fire"} & \multirow{7}{*}{Ed Sheeran} & \multirow{2}{*}{High Hopes} & \multirow{2}{*}{Bruce ...} \\
\cline{1-2}
4 & 24 Jan 2014 & & & & \\
\cline{1-2}\cline{5-6}
5 & 31 Jan 2014 & & & True & Avicii \\
\cline{1-2}\cline{5-6}
6 & 7 Feb 2014 & & & \multirow{2}{*}{\makecell{Christer Sjögren \\ sjunger Sinatra}} & \multirow{2}{*}{Christer Sjögren} \\
\cline{1-2}
7 & 14 Feb 2014 & & & & \\
\cline{1-2}\cline{5-6}
8 & 21 Feb 2014 & & & \multirow{3}{*}{True} & \multirow{3}{*}{Avicii} \\
\cline{1-2}
9 & 28 Feb 2014 & & & & \\
\cline{1-4}
... & ... & ... & ... &  & \\
\hline
31 & 31 July 2014 & \multirow{3}{*}{Prayer in C...} &	 \multirow{3}{*}{Lilly Wood \&...} & X & Ed Sheeran \\
\cline{1-2}\cline{5-6}
32 & 7 Aug 2014 &  &  & Honky Tonk Rebels &	Lasse Stefanz \\
\cline{1-2}\cline{5-6}
... & ... & & & ... & ... \\
\hline
42 & 16 Oct 2014 & \multirow{3}{*}{"The Days"} & \multirow{3}{*}{Avicii} & \multirow{2}{*}{X} & \multirow{2}{*}{Ed Sheeran} \\
\cline{1-2}
43 & 23 Oct 2014 & & & & \\
\cline{1-2}\cline{5-6}
44 & 30 Oct 2014	& & &  \multirow{2}{*}{Songs for Daddy} &  \multirow{2}{*}{Jill Johnson} \\
\cline{1-4}
... & ... & ... & ... &  &  \\
\hline
\end{tabular}
\end{adjustbox}
\\
\textbf{Adaper-tune:} I found a table of the year 2014.  It shows the week of the week, the song title, the artist, the title of the song, the album title, and the performer.  There are 11 times that Ed Sheeran was listed as the performer in the year.  The song title is "Timber" and the album is True. \\

\textbf{Fine-tune:} I found a table of Ed Sheeran's year in 2014. He was listed as the performer 14 times in the year 2014. The first time he was listed was on 3 January 2014 with the song "Timber" and the last time was on 4 April 2014 with "I See Fire". \\
\hline
\end{tabularx}
\end{adjustbox}
\caption{Example from the Tablesum dataset.}
\label{tab:tablesum_example}
\end{table*}

\begin{table}[t!]
\centering
\begin{tabular}{ccr}
\toprule
\multicolumn{2}{c}{\textbf{Adapter-tune}}  & 
  \multicolumn{1}{c}{\multirow{4}{*}{\textbf{\makecell{\#Trainable \\ parameters}}}} \\
\cmidrule{1-2} 
\textbf{\makecell{Encoder \\ adapters \\ removed}} & \textbf{\makecell{Decoder \\ adapters \\ removed}} & \\
\midrule
--    & --     & $6,343,680$ $(1.56\%)$ \\
 0--2 & 12--14 & $4,757,760$ $(1.17\%)$ \\
 0--4 & 12--16 & $3,700,480$ $(0.91\%)$ \\
 0--6 & 12--18 & $2,643,200$ $(0.65\%)$ \\
 0--8 & 12--20 & $1,585,920$ $(0.39\%)$ \\
 0--10 & 12--22 & $528,640$ $(0.13\%)$ \\
 0--11 & 12--22 & $264,320$ $(0.07\%)$ \\
 \midrule
 \multicolumn{2}{l}{\textbf{fine-tune}}  & $406,291,456$ $(100\%)$ \\
 \bottomrule
\end{tabular}
\caption{Trainable parameters in the encoder and decoder. Encoder adapter layers are numbered from 0--11 and decoder adapter layers are numbered from 12--22. $x$--$y$ implies all adapter layers from $x$ to $y$ inclusive.}
\label{tab:parameters}
\end{table}

\begin{figure*}
\begin{subfigure}{.33\textwidth}
  \centering
  \includegraphics[width=1.\linewidth]{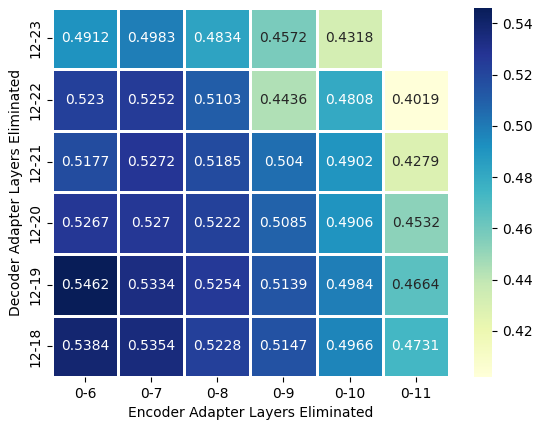}
  \caption{FeTaQA Rouge-L scores}
  \label{fig:rougel_fetaqa}
\end{subfigure}
\begin{subfigure}{.33\textwidth}
  \centering
  \includegraphics[width=1.\linewidth]{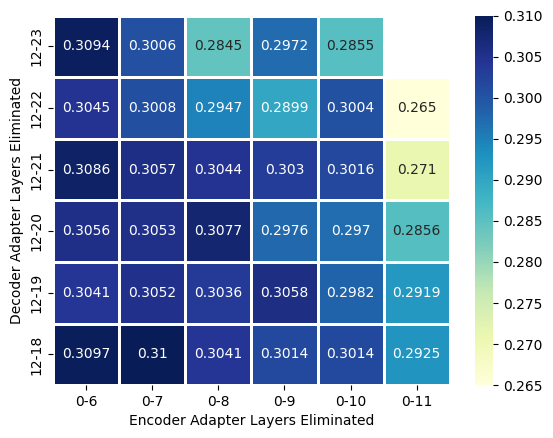}
  \caption{Tablesum Rouge-L scores}
  \label{fig:rougel_tablesum}
\end{subfigure}
\begin{subfigure}{.33\textwidth}
  \centering
  \includegraphics[width=1.\linewidth]{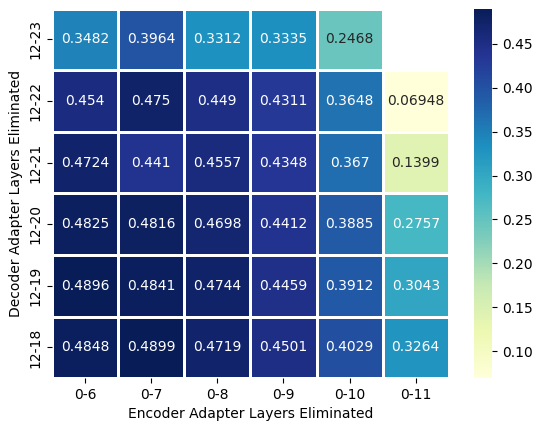}
  \caption{NarrativeQA Rouge-L scores}
  \label{fig:rougel_narrativeqa}
\end{subfigure}
\newline

\begin{subfigure}{.33\textwidth}
  \centering
  \includegraphics[width=1.\linewidth]{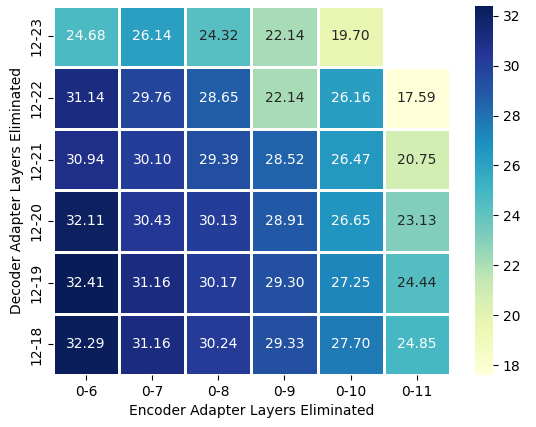}
  \caption{FeTaQA sacreBLEU scores}
  \label{fig:bleu_fetaqa}
\end{subfigure}
\begin{subfigure}{.33\textwidth}
  \centering
  \includegraphics[width=1.\linewidth]{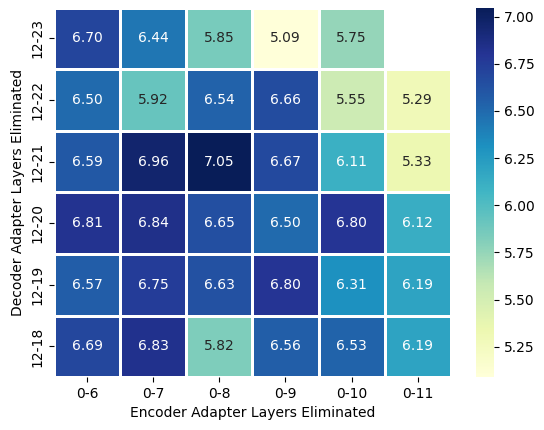}
  \caption{Tablesum sacreBLEU scores}
  \label{fig:bleu_tablesum}
\end{subfigure}
\begin{subfigure}{.33\textwidth}
  \centering
  \includegraphics[width=1.\linewidth]{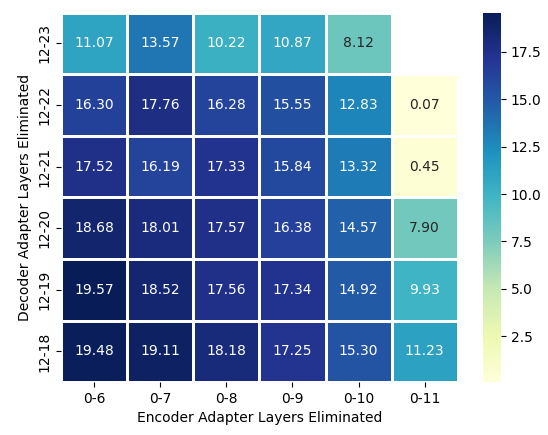}
  \caption{NarrativeQA sacreBLEU scores}
  \label{fig:bleu_narrativeqa}
\end{subfigure}
\caption{Adapter layer ablation scores. The X-axis represents range of encoder adapter layers deleted, the Y-Axis represents range of decoder adapter layers deleted. $x$-$y$ implies all adapter layers from $x$ to $y$ inclusive. There are 36 model ablation configurations displayed. The ablation starts from $0$ to $6$ encoder adapter layers removal and 12 to 18 decoder adapter layer removal represented by the bottom left cell ((0--6), (12--18)) and progressively increases deletion of encoder adapter layers along the X-axis and decoder adapter layers along the Y-axis.}
\label{fig:encoder-decodeAblation}
\end{figure*}

\section{Results}
\label{sec:results}
 We compare the results of our baseline fine-tuned models with the state-of-the-art fine-tuned models in Section \ref{sec:result_fine_tune}. We address (RQ1) ``How does adapter-tuning perform compared to fine-tuning in the context of multi-modal input?'' in Section \ref{sec:result_adapter_tune} and (RQ2) ``Do all adapter
layers across the encoder and decoder contribute equally to performance across tasks/modalities?'' in \ref{sec:result_ablation}.

\subsection{Fine-Tuned Models}
\label{sec:result_fine_tune}
We study the results of our baseline fine-tuned models with the state-of-the-art fine-tuned models for the 3 datasets.  The results of the experiments are shown in Table~\ref{tab:tablesum_results}. We observe that for the Tablesum dataset, our fine-tuned model outperform the best state-of-art T5 model on Rouge-1 by $3.8\%$ , Rouge-2 by $4.3\%$ and Rouge-L score by $4\%$. This can be attributed to fine-tuning our model on the clean version of the dataset.  Our fine-tuned models perform comparably to the state-of-the-art T5-large on FeTaQA dataset, i.e, $0.2\%$ on Rouge-1, $0.01\%$ higher on Rouge-2, and $0.04\%$ higher on Rouge-L. Our fine-tuning results on NarrativeQA are lower than state-of-the-art models trained with sophisticated reasoning architecture. The focus of this work was primarily on comparing fine-tuning and adapter-tuning and hence we leave explicit reasoning as part of future work. 

\subsection{Adapter-Tuned Models}
\label{sec:result_adapter_tune}
We address (RQ1) by comparing the performance of adapter-tuned models to our baseline fine-tuned models. For Tablesum, as observed in Table~\ref{tab:tablesum_results} fine-tuning(baseline) marginally outperforms adapter-tuning with $0.7\%$ higher Rouge-1 and  $0.4\%$ higher Rouge-L scores while having the same Rouge-2 score. For FeTaQA, adapter-tune shows a larger performance gain with $1.9\%$ on Rouge-1 and Rouge-L and $2.1\%$ on Rouge-2 compared to fine-tuning. The insignificant gains of fine-tuning over adapter-tuning in tabular QA can be attributed to catastrophic forgetting  \citep{FRENCH1999128,kirkpatrick2017overcoming,chen-etal-2020-recall} induced by differences in the distribution of downstream tabular data format from the original text data format of pre-training.

To explore this phenomenon further, we analyse examples from FeTaQA dataset in Table~\ref{tab:fetaqa_examples} where adapter-tuning outperforms fine-tuning. We observe that the fine-tuned model is unable to disambiguate surface-form similarities from the column semantics in the first example. The intended semantics of the named-entity \emph{Akhila Kishore} in the question is \emph{Actor}. While the surface-form is similar to the column value \emph{Akhila}, the intended semantics is that of the column header \emph{Role}. The fine-tuned model wrongly predicts the second and third row of the tabular context as correct grounding of information while adapter-tuning is able to disambiguate and predicts information from the first 2 rows as answer. We observe that the fine-tuned model also predicts information from the wrong column \emph{Director} instead of \emph{Cast} in the second example. Adapter-tune correctly identifies the column but partially generates the required information in the prediction. The third example depicts both non-factual and non-fluent prediction by the fine-tuned model.

We demonstrate an example of a hierarchical table of Tablesum in Table~\ref{tab:tablesum_example}. The question requires aggregation on the table cells and has various summary-like targets associated with it. The hierarchical table mentions \emph{Ed Sheeran} $3$ times, but the actual number of occurrence is $10$ times, from \emph{Week 3} to \emph{Week 9}, \emph{Week 31} and from \emph{Week 42} to \emph{Week 43}. Our table transformation process handles this to produce a regular table with $10$ cells containing \emph{Ed Sheeran} as value. The models can simply aggregate over the mentions. As shown in Table~\ref{tab:tablesum_example}, both models generates long answers summarizing information from the context table. However, as the models do not explicitly handle cell aggregation, we observe factual mistakes in both adapter-tuned and fine-tuned models. The models find Tablesum samples challenging even though the generated language is fluent and readable. 

For textual QA, on the NarrativeQA dataset, adapter-tuning performs comparable to fine-tuning with the adapter-tuned model achieving $0.8\%$ lower Rouge-1, $1.8\%$ higher  Rouge-2 and $1.5\%$ lower Rouge-L scores than fine-tuning. 

We conclude that adapter-tuning performs better than fine-tuning for out-of-domain tabular data and comparable performance on in-domain text.

\begin{figure}[ht!]
 \centering
   \includegraphics[width=\columnwidth]{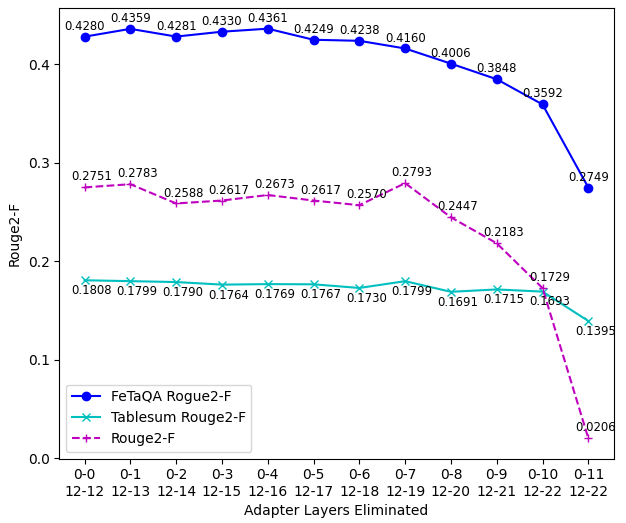}
 \caption{Adapter layer ablation Rouge2 F-scores. The X-axis depicts encoder-adapter layers (0--11) and decoder adapter layers (12--23) deleted progressively. Each $(x-y) \atop  (r-s)$ represents F-score with encoder layers $p$ to $q$ deleted and decoder layers $r$ to $s$ deleted.}
\label{fig:rouge2_ablation_all}
\end{figure} 

\begin{figure}[ht!]
 \centering
   \includegraphics[width=\columnwidth]{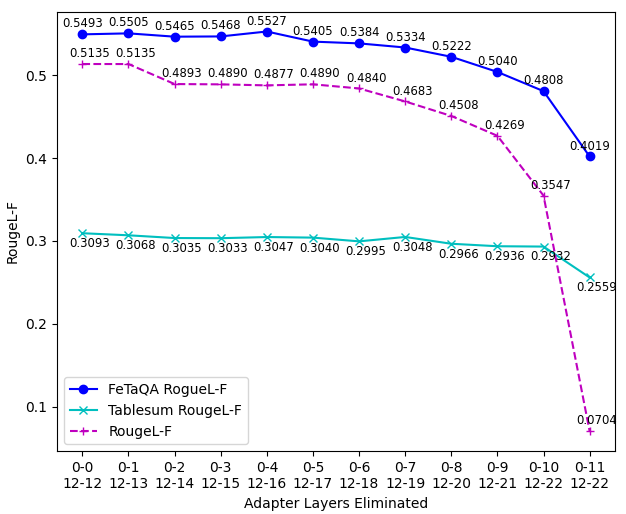}
 \caption{Adapter layer ablation Rouge-L scores. The X-axis depicts encoder-adapter layers (0--11) and decoder adapter layers (12--23) deleted progressively. Each $(x-y) \atop  (r-s)$ represents F-score with encoder layers $p$ to $q$ deleted and decoder layers $r$ to $s$ deleted.}
\label{fig:rogueL_ablation_all}
 \label{fig:rougel_all}
\end{figure} 

\begin{figure}[ht!]
 \centering
   \includegraphics[width=\columnwidth]{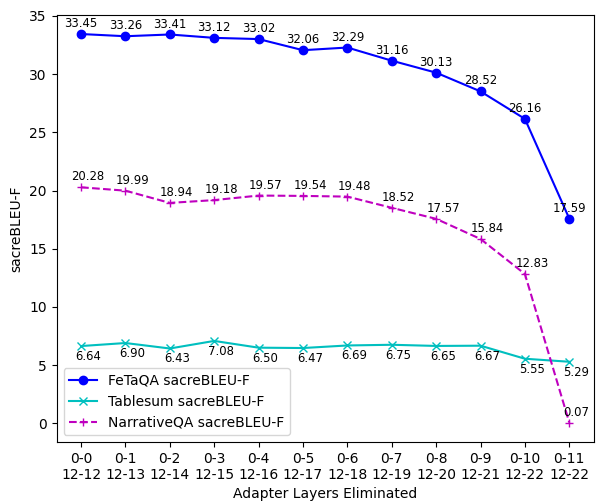}
 \caption{Adapter layer ablation sacreBLEU F-scores.  The X-axis depicts encoder-adapter layers (0--11) and decoder adapter layers (12--23) deleted progressively. Each $(x-y) \atop  (r-s)$ represents F-score with encoder layers $p$ to $q$ deleted and decoder layers $r$ to $s$ deleted.}
\label{fig:bleu_ablation_all}
 \label{fig:bleu_all}
\end{figure}  

\subsection{Ablation of adapter layers}
\label{sec:result_ablation}
We study (RQ2) by ablating adapter layers in both the encoder and decoder modules. We uniformly eliminate successive adapter layers from both encoder and decoder starting from the first layer in both modules and finally deleting all layers. This leads to 12 experiments corresponding to 12 encoder and 12 decoder adapter layers. We number the encoder adapter layers from 0--11 and the decoder adapter layers from 12--23. We measure the performance of the models using Rouge-2, Rouge-L\footnote{\url{https://pypi.org/project/rouge-score/}} and sacreBLEU\footnote{\url{https://github.com/mjpost/sacreBLEU}} scores. The F-scores for each dataset (NarrativeQA, Tablesum, FeTaQA) are shown in Figure~\ref{fig:rouge2_ablation_all}, ~\ref{fig:rogueL_ablation_all} and~\ref{fig:bleu_ablation_all}, respectively. We observe that as more adapter layers are eliminated, the performance drops across all datasets. However, the performance drop is minimal until the last adapter layers are also deleted. The inflection point varies across dataset but is limited to the last $2$ layers of the encoder and decoder. For the NarrativeQA dataset, this point is when all layers till the second last adapter layer from both the encoder and decoder are deleted. For the FeTaQA and Tablesum datasets, the performance drops sharply only when the last encoder and decoder layers are removed. 

To analyze contribution of the $i$-{th} adapter layer of encoder and decoder to performance, we perform ablation of adapter layers (0--6), (0--7), \ldots, (0--11) from encoder and adapter layers (12--18), (12--19), \ldots, (12--23) from decoder (decoder layers are numbered 12--23). This leads to $36$ configurations where a configuration ($p$--$q$, $r$--$s$) represents removal of all encoder adapters from $p$-th to $q$-th layer and all decoder adapters from $r$-th to  $s$-th. The results are shown in Figure~\ref{fig:encoder-decodeAblation}. We observe that performance remains comparable as we progressively eliminate adapter layers from encoder and decoder until the last layers. The performance drops steeply when we remove the last encoder and decoder adapter layers depicted towards the top-right corner of RougeL scores in Figures~\ref{fig:rougel_fetaqa},~\ref{fig:rougel_tablesum},  and~\ref{fig:rougel_narrativeqa} and BLEU scores in Figures~\ref{fig:bleu_fetaqa},~\ref{fig:bleu_tablesum},  and~\ref{fig:bleu_narrativeqa}. 
This implies that last adapter layers learns most of the domain information.

We also observe that the last encoder and decoder layers contribute differently to performance. Removing the last encoder layer (column 0--11) leads to substantial score drop across all decoder layers. This indicates that the last encoder layer is indispensable. Keeping only the last decoder adapter (row 12--23) is comparable to keeping last two last encoder layers (column 0--10). We also observe that retaining just the last $50\%$ of adapter layers from both encoder and decoder increases parameter efficiency by $0.7\%$ parameters as summarized in Table~\ref{tab:parameters} without significant compromise to performance. 

\section{Conclusion}
We are the first to study parameter-efficient transfer learning over tables and text for abstractive question answering using adapters. We demonstrate that parameter efficient adapter-tuning outperforms fine-tuning on out-of-domain tabular data and achieves comparable results on in-domain textual data. 

We propose a transformation from hierarchical tables to regular ones and further into a sequential form compatible with pre-trained model. We extend an existing ablation study of adapter layers to encoder-decoder setting and demonstrate that adapter layers from the end of the encoder is indispensable to encoding modality specific information than decoder adapter layers at the same level. 

Our results are useful for exploring scalability of QA models in memory constrained situations with comparable performance while scaling across modalities using light-weight adapters.

One of the limitations of our work is that our models do not explicitly reason and aggregate over the table cells. This might lead to fluent but factually incorrect answers on challenging Tablesum dataset. Addressing this limitation is left as future work.

\section{Acknowledgements}
We would like to thank Elsevier for their support throughout this project and funding this work. This work was also supported by
the NWO Innovational Research Incentives Scheme Vidi (016.Vidi.189.039),
the NWO Smart Culture - Big Data / Digital Humanities (314-99-301),
the H2020-EU.3.4. - SOCIETAL CHALLENGES - Smart, Green And Integrated Transport (814961).
All content represents the opinion of the authors, which is not necessarily shared or endorsed by their respective employers and/or sponsors.

\clearpage
\bibliographystyle{acl_natbib}
\bibliography{references}

\clearpage
\appendix
\section*{APPENDICES}
We provide further details on statistics of the datasets used (Appendix~\ref{sec:dataset_statistics}) and on the Rouge-2 scores for an encoder-decoder adapter layer ablation study (Appendix~\ref{appendix:section:adapter-ablation-rouge}).

\section{Dataset Statistics}
\label{sec:dataset_statistics}
Statistics of the three datasets, i.e., Tablesum, FeTaQA and NarrativeQA are listed in Table \ref{tab:dataset_statistics}. Tablesum has the longest answer length. The answers are summary-like, often, describing aspects of the table contents. The FeTaQA dataset contains answers of mostly single sentences and targeted towards specific facts asked in the question. The NarrativeQA dataset focuses on questions from stories. The answer lengths vary from single words to long sentences. For the tabularQA dataset, Tablesum contains larger tables than the FeTaQA dataset even though it is limited to $200$ unique tables over which questions are asked. The FeTaQA dataset's tables contain more columns on average than Tablesum.  

\medskip
\begin{center}
\bottomcaption{Dataset Statistics}
\begin{supertabular}{lr}
\toprule
\multicolumn{2}{c}{\textbf{Tablesum}} \\
\midrule
 Domain & Open  \\
 Modality & Table \\
 Table-type & Regular \\
 Training samples & 798 \\
 Validation samples & 200 \\
 Test samples & -- \\
 Max question length & 114 \\
 Max target length & $1,579$ \\
 Max table row & 155 \\
 Max table column & 8 \\
 \midrule
 \multicolumn{2}{c}{\textbf{FeTaQA}} \\
 \midrule
  Domain & Open  \\
 Modality & Table \\
 Table-type & Hybrid \\
 Training samples & $7,326$ \\
 Validation samples & $1,001$ \\
 Test samples & $2,003$ \\
 Train max question length & 165 \\
 Train max target length & 338 \\
 Train max table rows & 34 \\
 Train max table columns & 30 \\
 Val max question length & 182 \\
 Val target length & 325 \\
 Val max table rows & 34 \\
 Val max table columns & 22 \\
 Test max  question length & 193 \\
 Test max target length & 295 \\
 Test max table lows & 34 \\
 Test max table columns & 22 \\
 \midrule
 \multicolumn{2}{c}{\textbf{NarrativeQA}} \\
\midrule
  Domain & Stories \\
  Modality & Text \\
 Training samples & $65,494$ \\
 Validation samples & $6,922$ \\
 Test samples & $21,114$ \\
 Train max question length & 175 \\
 Train max target length & 171 \\
 Train max context length & $6,045$ \\
 Val max question length & 158 \\
 Val target length & 187 \\
 Val max context length & $6,033$ \\
 Test max question length & $1,220$ \\
 Test target length & 224 \\
 Test max context length & $6,090$ \\
\bottomrule
\end{supertabular}
\label{tab:dataset_statistics}
\end{center}

\section{Encoder-Decoder Adapter Layer Ablation Rouge-2 Scores}
\label{appendix:section:adapter-ablation-rouge}

\begin{figure*}[t!]
\begin{subfigure}{.33\textwidth}
  \centering
  \includegraphics[width=1.\linewidth]{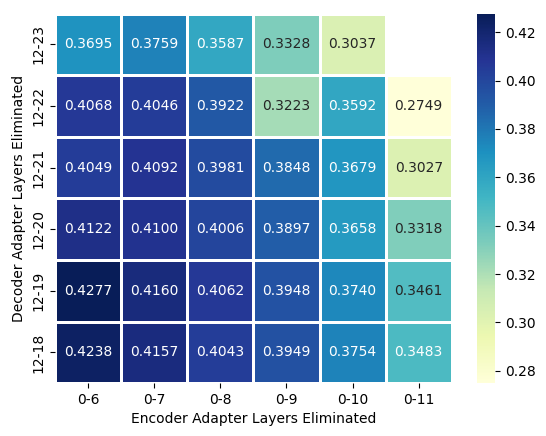}
  \caption{FeTaQA Rouge-L scores}
  \label{fig:appendix_fetaqa_rouge2}
\end{subfigure}
\begin{subfigure}{.33\textwidth}
  \centering
  \includegraphics[width=1.\linewidth]{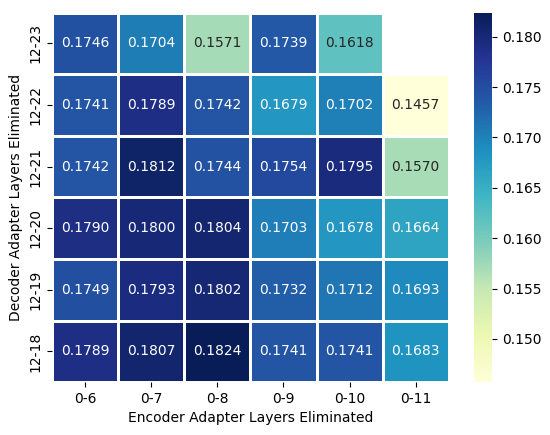}
  \caption{Tablesum Rouge-L scores}
   \label{fig:appendix_tablesum_rouge2}
\end{subfigure}
\begin{subfigure}{.33\textwidth}
  \centering
  \includegraphics[width=1.\linewidth]{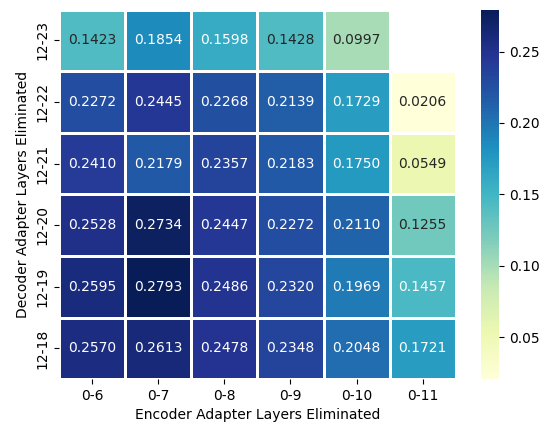}
  \caption{NarrativeQA Rouge-L scores}
  \label{fig:appendix_narrativeqa_rouge2}
\end{subfigure}
\caption{Adapter layer Rouge-2 ablation scores. The X-axis represents range of encoder adapter layers deleted, the Y-Axis represents range of decoder adapter layers deleted. $x$-$y$ implies all adapter layers from $x$ to $y$ inclusive. There are 36 model ablation configurations displayed. The ablation starts from $0$ to $6$ encoder adapter layers removal and 12 to 18 decoder adapter layer removal represented by the bottom left cell ((0--6), (12--18)) and progressively increases deletion of encoder adapter layers along the X-axis and decoder adapter layers along the Y-axis.}
\label{fig:appendix_rouge2}
\end{figure*}

Ablation results (Rouge-2 F-scores) of $36$ configurations of adapter layers deleted from the later half of the encoder and decoder. Deleting the last encoder adapter layers leads to massive drop in performance as observed in the last three columns of Figures \ref{fig:appendix_fetaqa_rouge2}, \ref{fig:appendix_tablesum_rouge2} and \ref{fig:appendix_narrativeqa_rouge2}. However, deleting the last decoder adapter layers results in better performance in comparison to the encoder layers at the same level as observed from the top 3 rows.

\vfill

\vspace*{15cm}

\vfill

\end{document}